\documentclass[11pt]{article}

\usepackage[]{emnlp2023}

\usepackage{float}
\usepackage{times}
\usepackage{latexsym}
\usepackage{multirow}
\usepackage{amsmath}
\usepackage{mathtools}
\usepackage{relsize}
\usepackage{natbib}

\usepackage[T1]{fontenc}

\usepackage[utf8]{inputenc}

\usepackage{microtype}

\usepackage{microtype}
\usepackage{graphicx}
\usepackage{subfigure}
\usepackage{booktabs} 
\usepackage{url}
\usepackage{amsmath}
\usepackage{array}
\usepackage{adjustbox}
\usepackage{multirow}
\usepackage{longtable}
\usepackage{enumitem}

\usepackage[T1]{fontenc}

\usepackage[utf8]{inputenc}

\usepackage{microtype}

\usepackage{inconsolata}

\usepackage{graphicx}

\title{\metric: \metricexp}

\author{Nirupan Ananthamurugan, Dat Duong, Philip George, Ankita Gupta \\
  University of Massachusetts Amherst, Amherst, MA, USA \\
  \texttt{\{nananthamuru,dduong,pgeorge,ankitagupta\}@umass.edu} \\\\
  \textbf{Sandeep Tata, Beliz Gunel} \\
  Google Research, Mountain View, CA, USA \\
  \texttt{\{tata,bgunel\}@google.com} \\}

\newcommand{\metric}{{\bf CASPR}}
\newcommand{\metricexp}{{\bf Automated Evaluation Metric for Contrastive Summarization}}
\begin{document}
\maketitle
\begin{abstract}

Summarizing comparative opinions about entities (e.g., hotels, phones) from a set of source reviews, often referred to as \textit{contrastive summarization}, can considerably aid users in decision making. However, reliably measuring the \textit{contrastiveness} of the output summaries without relying on human evaluations remains an open problem. Prior work has proposed token-overlap based metrics, Distinctiveness Score \citep{iso-etal-2022-comparative}, to measure contrast which does not take into account the sensitivity to meaning-preserving lexical variations. In this work, we propose an automated evaluation metric $\metric$  to better measure contrast between a pair of summaries. Our metric is based on a simple and light-weight method that leverages natural language inference (NLI) task to measure contrast by segmenting reviews into single-claim sentences and carefully aggregating NLI scores between them to come up with a summary-level score. We compare $\metric$ with Distinctiveness Score and a simple yet powerful baseline based on BERTScore. Our results on a prior dataset \textsc{CoCoTRIP} \citep{iso-etal-2022-comparative} demonstrate that $\metric$ can more reliably capture the \textit{contrastiveness} of the summary pairs compared to the baselines.

\end{abstract}

\section{Introduction}

\begin{figure}[t]
\centering
\includegraphics[width=
\linewidth]{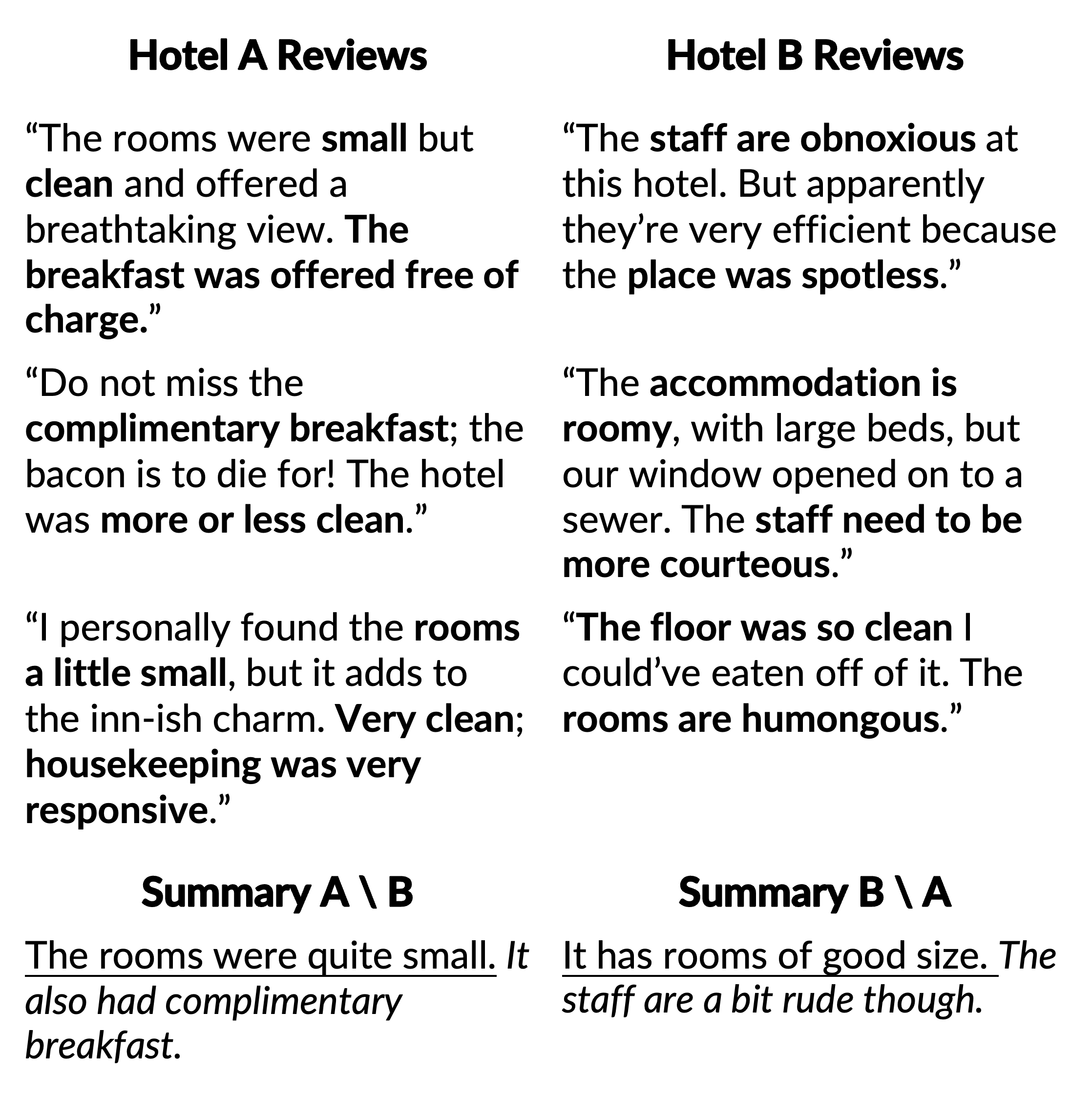}
\caption{Contrastive summaries for Hotel A and Hotel B, denoted as $A \setminus B$ and $B \setminus A$, respectively. The \underline{underlined} sentences in the summaries assign \textbf{different values} ('small', 'good size') to the \textbf{same aspect} (room size), and are therefore contrastive. The \textit{italicized} sentences in the summaries describe \textbf{different aspects} (breakfast and staff) -- also highlighting differences to help users decide.}
\label{fig:data_general}
\end{figure}

Consider the familiar problem of using the web to compare two entities A and B while shopping for products ("Which phone better suits my needs?"), planning travel ("Which hotel should I book for my anniversary trip?"), or even looking at universities ("Which university is better for my career?"). Parsing subjective information from multiple reviews to identify \textit{contrastive opinions} ("the room had modern decor" vs "the room's decor was outdated") is time-consuming. Therefore, the problem of generating summaries containing comparative opinions of two entities, or \textit{contrastive summaries}, is of practical importance.

Given reviews of two entities A and B, the task of \textbf{contrastive summarization} \cite{iso-etal-2022-comparative} involves generating comparative summaries conditioned on both entities, and containing no overlapping information. For each pair of entities $A$ and $B$, summary $A \setminus B$ contains opinions in A that contrast with or are not present in B. This is illustrated in Figure~\ref{fig:data_general}.

However, there currently exists no reliable automated evaluation metric to measure this contrast, which is an obstacle to further work on this task without conducting large-scale human evaluations. Distinctiveness Score (DS) \cite{iso-etal-2022-comparative} treats token overlap between comparative summaries as a proxy for contrast. In our work, we demonstrate that this approach can have considerable limitations. We propose an automated evaluation metric, $\metric$, to measure contrast between a pair of summaries obtained via the contrastive summarization task. $\metric$ evaluates the logical relationships between pairs of summaries after decomposing summaries into single-claim sentences, employing an out-of-the-box NLI model to compute scores between each pair of sentences from the two summaries, and carefully aggregating scores to come up with a single \textit{contrastiveness} score for a pair of summaries. We demonstrate that $\metric$ can capture logical contrast beyond lexical and semantic variations on the \textsc{CoCoTrip} dataset, unlike existing baselines. We discuss possible future lines of investigation in this direction and release our modified dataset.

\section{Approach}
In this section, we describe the baselines, DS \cite{iso-etal-2022-comparative} and $BS^{-1}$, and how our proposed metric $\metric$ measures logical contrast beyond lexical and semantic variations.

\subsection{Baselines}
\subsubsection{Distinctiveness Score}
Distinctiveness Score (DS) \cite{iso-etal-2022-comparative} was introduced to measure contrast between \textit{three} summaries - two contrastive summaries $A\setminus B$ and $B\setminus A$, and a \textit{common} summary $A \cap B$. 
$$
DS = 1 - \frac{\sum_{(y,z)\in \hat{\Upsilon}^{2}} \vert \mathcal{W}_{y} \cap \mathcal{W}_{z} \vert - 2 \vert \bigcap_{y \in \hat{\Upsilon}}W_y \vert}{\vert \bigcup_{y \in \hat{\Upsilon}}W_y \vert},
$$
Where $\hat{\Upsilon}:=\{A\setminus B, B \setminus A, A \cap B\}$, $W_y$ represents the token bag of summary $y \in \hat{\Upsilon}$, and $\hat{\Upsilon}^{(2)}$ represents 2-subsets of $\hat{\Upsilon}$.
$A \cap B$ contains opinions that are common between the entities A and B. For simplicity, we only consider the two contrastive summaries and hence modify the definition of the DS to remove the common summary from the set:
$$
DS = 1 - \frac{\vert \mathcal{W}_{A \setminus B} \cap \mathcal{W}_{B \setminus A} \vert}{\vert \mathcal{W}_{A \setminus B} \cup \mathcal{W}_{B \setminus A} \vert},
$$
The original DS simply measures the lexical overlap without differentiating between contrastive and common summaries; in fact, it can be extended to multiple entities and summaries. Therefore, removing the common summary from the equation does not alter the calculation or the definition of the DS.
The lower the token overlap, the higher the value of DS. It is scaled to be within [0,100] with higher numbers indicating higher contrast (low lexical overlap). As illustrated in Table~\ref{table:1}, the reliance on token overlap leads to an erroneous DS score in cases where text is compared to its paraphrase.

\begin{table}[h!]
\centering
\small
\begin{tabular}{c c c  c } 

 (i) & \multicolumn{3}{l}{ A: The hotel is sparkly clean.} \\
 & \multicolumn{3}{l}{ B: The hotel was kept very tidy.} \\
 \\
 (ii) & \multicolumn{3}{l}{A: The hotel is clean.}\\
 & \multicolumn{3}{l}{ B: The hotel is not clean}\\
\\
& \multicolumn{1}{|c|}{DS} & \multicolumn{1}{c|}{$BS^{-1}$}  & \metric \\ 
\hline
(i) & \multicolumn{1}{|c|}{78} & \multicolumn{1}{c|}{25} & 0 \\
\hline
(ii) & \multicolumn{1}{|c|}{20}  & \multicolumn{1}{c|}{26} &  100\\

 \hline
\end{tabular}
\caption{\textbf{Drawbacks of $DS$ \cite{iso-etal-2022-comparative} and $BS^{-1}$: } (i) $DS$ assigns high contrast (78) to semantically similar sentences due to low lexical overlap, and (ii) $DS$ (20) and $BS^{-1}$ (26) both fail to detect the logically-induced high contrast. $\metric$ behaves as expected in both cases.}
\label{table:1}
\end{table}

\subsubsection{Inverted BERTscore $(BS^{-1})$}
We include a simple but powerful baseline that we refer to as inverted BERTScore ($BS^{-1}$). BERTScore was introduced \cite{zhang2019bertscore} to measure semantic similarity by aggregating cosine distances between contextual embeddings of generated and reference sequences. We use the HuggingFace implementation\footnote{\url{https://huggingface.co/spaces/evaluate-metric/bertscore}} which outputs a similarity score in the $[0,1]$ range. We use the F1 score of BERTScore as the scores are symmetrical with order of the inputs. We invert and rescale this to get $BS^{-1} = 100* (1 - \text{BERTScore})$ as a measure of contrastiveness, with higher values indicating higher contrast. We compute $BS^{-1}$ on summaries $S^{A \setminus B}, S^{B \setminus A}$ as follows,
\begin{equation*}
{
BS^{-1} = 1 - BS(S^{A \setminus B}, S^{B \setminus A})}
\end{equation*}


\noindent $BS^{-1}$ is in the range [0,100] with higher numbers indicating higher contrast. $BS^{-1}$ is more robust to lexical changes than DS due to its use of BERT word embeddings, as evidenced by our example in Table~\ref{table:1}. However, it is not sensitive to sentence-level logical relationships; concretely, as in Table~\ref{table:1}, it fails to detect that the addition of a single token "not" has created a logically contradictory sentence, and therefore assigns a low contrast score of $26$.



\subsection{\metric}
\label{sec:nli_approach}
We propose a metric $\metric$ based on a method that evaluates the logical relationships between pairs of sentences, assigns a score to these logical relationships, and finally aggregates them across all sentence pairs in the two contrastive summaries.\\


\noindent \textbf{Natural Language Inference (NLI)} We use the textual entailment task, also referred to as NLI, to compare the logical relationship between two sentences. More formally, given a premise and a hypothesis, an NLI model determines the inference relationship (entailment, contradiction, neutral) between them. For each pair, we consider the entailment relationship in both directions. For our experiments we use off-the-shelf NLI model: \texttt{roberta-large-mnli}\footnote{\url{https://huggingface.co/roberta-large-mnli}} \cite{Liu2019RoBERTaAR} \\

\noindent \textbf{Forming Single-Claim Sentences} Our dataset contains long sentences embodying multiple claims, which is commonplace in reviews. An example complex-claim sentence would be "The hotel's breakfast is included in the room's price, but a little expensive." which we break down to "The hotel's breakfast is included in the room's price." and "The hotel's breakfast is a little expensive." This process helps us determine the fine-grained logical relationships between sentences with NLI models more reliably. We decompose the original sentences using \texttt{gpt-3.5-turbo} text generation model from OpenAI with few-shot prompting. Refer to Appendix~\ref{sec:sentence_decomposition} for the specific prompt we used, along with the output \textit{single-claim sentence}, as well as details on other hyperparameters.\\

\noindent \textbf{Scoring and Aggregation} For two contrastive summaries $S^{A\setminus B}, S^{B\setminus A}$, CASPR is computed as the average contrastiveness score across all constituent sentences in these summaries. To score a single constituent sentence $s_i^{A \setminus B}$ from summary $S^{A\setminus B}$, we compare $s_i^{A \setminus B}$ with all sentences from the other summary $s^{B \setminus A}$, and compute a contrastiveness score depending on its comparison-level labels with these sentences.

Consider a \textit{single comparison} of our sentence $s_i^{A \setminus B}$ with another sentence $s_j^{B \setminus A}$ from summary $S^{B\setminus A}$. To get a \textit{label} for this comparison, we evaluate the NLI label in both directions. The intuition for this is that NLI is \textit{directional}, since for the same pair of sentences, swapping the hypothesis and premise changes the relationship between them. Considering this, we assign the label for a single comparison as follows:
\begin{itemize}
    \item If both directions are `neutral', then the label for that comparison is `neutral'.
    \item If the labels are (`neutral', `contradiction') or (`contradiction', `neutral'), the label for this comparison is `contradiction'.
    \item If the labels are (`neutral', `entailment') or (`entailment', `neutral'), the label for this comparison is `entailment'.
    \item If the labels are (`contradiction', `entailment') or (`entailment', `contradiction'), the label for this comparison is `neutral'. This was a design choice based on our observation that there were very few such cases. In those few such cases, human examination revealed the model had incorrectly classified neutral statements as contradiction/entailment. 
\end{itemize}

We now have a single comparison label for $s_i^{A \setminus B}$ with $s_j^{B \setminus A}$. We repeat the above process across all comparisons between $s_i^{A \setminus B}$ and sentences from summary $S^{B\setminus A}$. We process this set of labels according to the following conditions, to get a single score in the range [0,1] for our sentence $s_i^{A \setminus B}$.
\begin{itemize}
    \item \textbf{Case 1: All Comparisons are Neutral} - If $s_i^{A \setminus B}$ was \textit{neutral} to all sentences in the other summary, we take this to mean that $s_i^{A \setminus B}$ mentions an aspect that is not mentioned in the other summary. We consider this a valid case of contrast, and assign it a score of +1.
\end{itemize}
If \textit{Case 1} is not true, then this means that there are some contradictions and/or entailments, i.e. there are some comparisons which are discussing the same aspect, and have values that agree/disagree with each other. In this case, we ignore the neutralities and \textit{count the number of contradictions and entailments}.
\begin{itemize}
    \item \textbf{Case 2: Number of Contradictions > Number of Entailments} - We take this to mean that the sentence is overall contrastive w.r.t the other summary, and score it +1.
    \item \textbf{Case 3: Number of Entailments >= Number of Contradictions} - We take this to mean that the sentence is overall similar w.r.t the other summary, and score it -1.
\end{itemize}
Note from the above, that in cases where the tally of entailments and contradictions is tied, we bias the score in favour of entailments; we found that the model was less sensitive to entailments in practice, and we decided to conservatively score ties in favor of similarity rather than contrast to offset this. We now have a +1/-1 score for our sentence $s_i^{A \setminus B}$. We now repeat this process for all sentences across both summaries $S^{A \setminus B}$ and $S^{B \setminus A}$ and take the \textit{mean score} to get CASPR.

\section{Experiments}
\subsection{Dataset: \textsc{CoCoTrip}}
We use the \textsc{CoCoTrip} dataset introduced by \cite{iso-etal-2022-comparative} which is designed to evaluate contrastive summarization methods. \textsc{CoCoTrip} consists of two contrastive summary sets $S^{A \setminus B}, S^{B \setminus A}$, for a pair of entities $A$ and $B$ summarized from review sets $R^A$ and $R^B$. The entities are hotels drawn from the TripAdvisor corpus \cite{10.1145/1835804.1835903}, with 8 reviews sampled per entity, for a total of 48 entity pairs. For each pair, they source reference summaries from three different annotators.
Note that \textsc{CoCoTrip} also provides one common summary for each entity pair which we do not consider as we are interested in evaluating \textit{contrastiveness}.

\subsection{Experimental Setup}
From the \textsc{CoCoTrip} dataset, we have reference summaries $S^{A \setminus B}, S^{B \setminus A}$ from three annotators for each entity pair. The summaries $S^{A \setminus B}_1, S^{B \setminus A}_1$ from the first annotator form our base dataset - the \textit{Reference Contrastive} dataset. Similarly, we pair reference summaries from the first and second annotators to obtain a dataset of summaries $S^{A \setminus B}_1, S^{A \setminus B}_2$ - the \textit{Reference Similar} dataset, as they are essentially summaries of the same entity by different annotators. We construct synthetic data with paraphrasing and logical negation that we describe below. We provide more details for each experiment along with examples from each constructed dataset in Appendix \ref{sec:appendix}.\\

\noindent \textbf{Logical Negations for \textit{Synthetic High Contrast}} To measure the ability of evaluation metrics to capture high contrast, we construct a synthetic dataset with logical negations. In particular, we manually negate all sentences in summary $S^{A \setminus B}_1$ to obtain a dataset consisting of summaries $S^{A \setminus B}_1, \neg S^{A \setminus B}_1$ for each entity pair. We refer to this dataset as \textit{Synthetic High Contrast} dataset. \\

\noindent \textbf{Self-Paraphrasing for \textit{Synthetic Low Contrast}} We construct a \textit{Synthetic Low Contrast} dataset by paraphrasing reference summaries $S^{A \setminus B}$. The resulting dataset consisting of summaries $S^{A \setminus B}$, $P(S^{A \setminus B})$ should exhibit low contrast. We paraphrase each sentence in the summary using OpenAI's \texttt{text-davinci-003}. Implementation details and examples are provided in Appendix \ref{sec:appendix}.\\

\noindent For an ideal metric, we expect the following: 
\begin{enumerate}
    \item Close to 0 score on \textit{Synthetic Low Contrast} dataset ($S^{A \setminus B}, P(S^{A \setminus B})$) as summaries are semantically equivalent.
    \item Close to 100 score on \textit{Synthetic High Contrast} dataset ($S^{A \setminus B}, \neg S^{A \setminus B}$) as summaries are logical contradictions of each other.
    \item Ranking from low to high scores to be \textit{Synthetic Low Contrast} ($S^{A \setminus B}, P(S^{A \setminus B})$), \textit{Reference Similar} ($S^{A \setminus B}_1, S^{A \setminus B}_2$), \textit{Reference Contrastive} ($S^{A \setminus B}, S^{B \setminus A}$), and \textit{Synthetic High Contrast} ($S^{A \setminus B}, \neg S^{A \setminus B}$) datasets with meaningful separation in scores.
\end{enumerate}

We publish our \textit{Synthetic Low Contrast} and \textit{Synthetic High Contrast} datasets in the \texttt{CASPR data repository}\footnote{\url{https://github.com/niru-umass-dev/caspr}}.



\subsection{Results}
\begin{table}[ht]
\centering
\small
\begin{tabular}{c | c | c } 

 Metric & $S^{A \setminus B},S^{B \setminus A}$ & $S^{A \setminus B}, \neg S^{A \setminus B}$\\ 
 \hline
 DS & 73.6 $\pm$ 0.9 & 44.5 $\pm$ 1.6\\ 

 BS$^{-1}$ & 72.5 $\pm$ 1.4 & 50.6 $\pm$ 1.7 \\

 \metric & 84.3$ \pm$ 3.8 & \textbf{98.1} $\pm$ 0.8\\

 \hline
\end{tabular}
\caption{Average scores over the \textsc{CoCoTrip} reference summaries $(S^{A\setminus B},S^{B \setminus A})$ and the \textit{Synthetic High Contrast} dataset $(S^{A \setminus B}, \neg S^{A \setminus B})$ that is designed for maximum contrast through logical negations. All scores are reported in the range $[0, 100]$. $\metric$ scores close to $100$ as desired, while DS \cite{iso-etal-2022-comparative} and $BS^{-1}$ actually register a \textbf{drop} in contrast scores.}
\label{table:2}
\end{table}

\begin{figure}[ht]
\centering
\includegraphics[scale = 0.42]{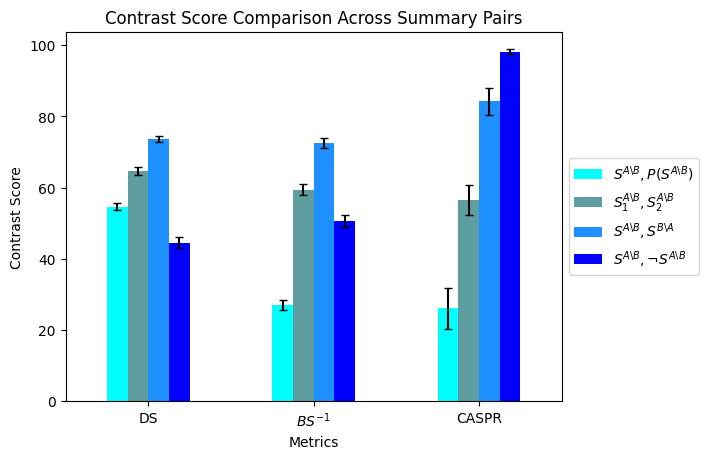}
\caption{Average contrast scores across \textit{Synthetic Low Contrast} ($S^{A \setminus B}, P(S^{A \setminus B})$), \textit{Reference Similar} ($S^{A \setminus B}_1, S^{A \setminus B}_2$), \textit{Reference Contrastive} ($S^{A \setminus B}, S^{B \setminus A}$), and \textit{Synthetic High Contrast} ($S^{A \setminus B}, \neg S^{A \setminus B}$) datasets. $\metric$ has a higher separation in contrast scores for all experiments, scores closest to 0 on \textit{Synthetic Low Contrast}, and scores closest to 100 on \textit{Synthetic High Contrast} as desired. }
\label{fig:result_plot}
\end{figure}

We use a non-parametric bootstrap~\cite[ch.~8]{wasserman2004all}
to infer confidence intervals for each method. We utilize $10^{4}$ bootstrap samples of summary pairs, to report 95$\%$ bootstrap confidence intervals (CI), via the normal interval method~\cite[ch.~8.3]{wasserman2004all}. In Table~\ref{table:2}, we see that all metrics register high scores on the Reference Contrastive dataset ($S^{A \setminus B}, S^{B \setminus A}$). On the Synthetic High Contrast dataset ($S^{A \setminus B}, \neg  S^{A \setminus B}$), $\metric$ \textit{increases} to $98.1$ and is a near-ideal score. However, both $DS(44.5)$ and $BS^{-1}(50.6)$ \textit{drop}, which supports our assertion that they are not sensitive to logical contrast. In Figure~\ref{fig:result_plot}, we see that $\metric$ moves from low to high scores as we traverse from Synthetic Low Contrast to Synthetic High Contrast, for which it has the \textit{closest} scores to 0 and 100 respectively. It has a meaningful separation of scores between all four datasets, since it can not only capture logical variations (where DS and $BS^{-1}$ clearly both fail) but can also capture semantic and lexical variations introduced by the other datasets.  $DS$ and $BS^{-1}$ have a much lower separation and counter-intuitively \textit{drop} for Synthetic High Contrast, compared to Reference Contrastive.
    
\section{Related Work}
Recently, there has been considerable interest in developing automated evaluation metrics in order to ensure faster and higher quality iteration on various ML tasks across different domains \cite{Adamson2021SSFDSF,Liu2023GEvalNE}. Few recent works \cite{10.5555/1620853.1620886,Gunel2023STRUMEA} develop contrastive summarization methods, but do not offer automated evaluation metrics. Most relevant to our work, collaborative decoding method \cite{iso-etal-2022-comparative} was proposed which contrasts token probability distributions to generate two contrastive summaries. They conduct automated evaluation of their summarization quality using ROUGE 1/2/L F1 scores \cite{Lin2004ROUGEAP}, BERTScore \cite{zhang2019bertscore}, and Distinctiveness Score. However, as illustrated in Table~\ref{table:1}, these metrics which are based on lexical overlap and BERTScore, do not capture simple logical relationship among sentence pairs (e.g., negations). Finally, NLI-based methods has been shown effective for evaluating \textit{factual consistency} of single entity summarization methods \cite{https://doi.org/10.48550/arxiv.2208.07316}. To the best of our knowledge, ours is the first work to utilize NLI-based methods for quantifying contrast among a pair of contrastive summaries.

\section{Conclusion}
In this paper, we propose an automated evaluation metric $\metric$ for the contrastive summarization task. $\metric$ evaluates the logical relationships between pairs of summaries after decomposing summaries into single-claim sentences, employing a textual entailment model, and carefully aggregating scores to come up with a summary-level \textit{contrastiveness} score. We demonstrate that $\metric$ can capture logical contrast beyond lexical and semantic variations on \textsc{CoCoTrip} dataset, unlike existing baselines. Regarding ease of implementation, our metric uses out-of-the-box models, hence does not need any fine-tuning.

\section{Limitations}
We list some of the limitations of our study which researchers and practitioners would hopefully benefit from when interpreting our results. We used \texttt{roberta-large-mnli} for NLI which is a small model, and \texttt{gpt-3.5-turbo} for forming single-claim sentences to demonstrate the benefits of using an approach based on NLI to measure contrast. However, we acknowledge that using state-of-the-art large language models such as \texttt{text-davinci-003} for both of these steps might yield much better results.
Also, we acknowledge that CASPR specifically focuses on measuring contrast among a pair of summaries. We consider our work to be complementary to other recent efforts in this research area, such as measuring factuality of summaries with respect to the source reviews  
\cite{laban-etal-2022-summac, glover2022revisiting} or measuring how well a metric captures the popular opinions or overall sentiment on various aspects mentioned in a summary \cite{brazinskas-etal-2020-unsupervised}.

\newpage
\bibliographystyle{plainnat} 
\bibliography{custom}

\newpage
\appendix
\section{Sentence Decomposition}
\label{sec:sentence_decomposition}
\begin{figure*}[t]
\includegraphics[width=\textwidth]{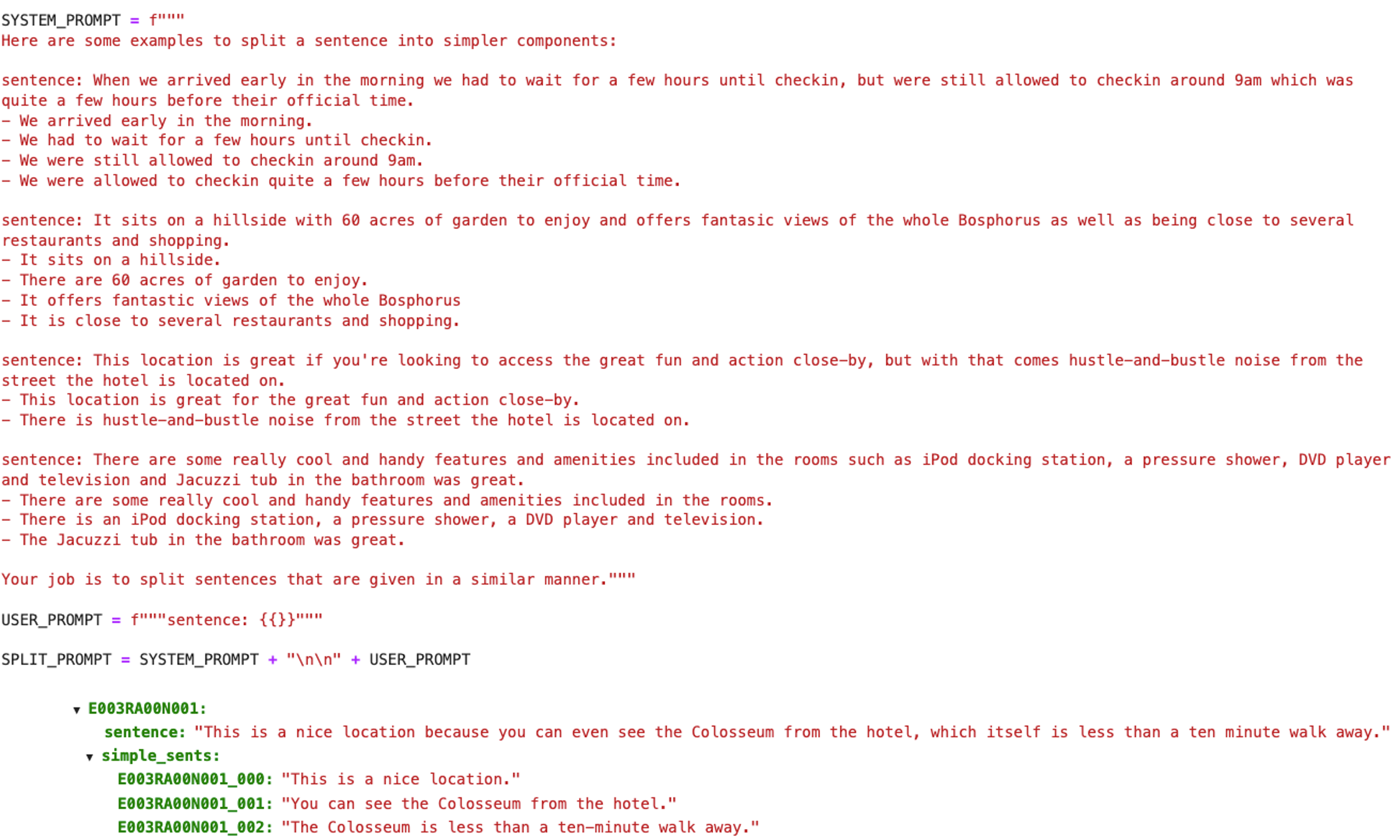}
\caption{Our System Prompt to \texttt{gpt-3.5-turbo} for the sentence splitting task. We also show an example from Reference Summary A of entity pair 3.}
\label{fig:sentence_split}
\end{figure*}

We used the \texttt{gpt-3.5-turbo} model from the OpenAI Chat Completion API. Figure~\ref{fig:sentence_split} shows the specific few-shot prompt we used, along with an example of an input \textit{complex-claim sentence} and output \textit{simple-claim sentences}. Table~\ref{table:experimental_datasets} shows an example summary $S_1^{A \setminus B}$ from the CoCoTrip dataset and the corresponding \textit{decomposed} summary formed of \textbf{single-claim sentences}.
\par We set \texttt{max\_tokens=256} and \texttt{temperature=0.5} to encourage the model to stick close to the structure of the original sentence. Note that although \texttt{gpt-3.5-turbo} is a chat-completion model\footnote{https://platform.openai.com/docs/models/gpt-3-5}, we treated every complex sentence as a new chat, and included only the current sentence in each API call without appending prior input and responses.

\section{Experimental Datasets}
\label{sec:appendix}

\begin{table*}[ht]
    \centering
    \small
    \begin{tabular}{|c|c|c|}
        \hline
         \textbf{Data Property} & \textbf{Data Type} & \textbf{Comparison} \\
        \hline
         Reference Contrastive & Reference & $S_1^{A \setminus B}$ vs. $S_1^{B \setminus A}$\\
        \hline
        Reference Similar & Reference & $S_1^{A \setminus B}$ vs. $S_2^{A \setminus B}$\\
        \hline
        Synthetic Low Contrast & Reference \& Paraphrased & $S_1^{A \setminus B}$ vs. $P(S_1^{A \setminus B})$ \\
        \hline
         Synthetic High Contrast & Reference \& Negated & $S_1^{A \setminus B}$ vs. $ \neg S_1^{A \setminus B}$ \\ 
        \hline
         Synthetic Contrast & Paraphrased & $P(S_1^{A \setminus B})$ vs. $P(S_1^{B \setminus A})$\\
         \hline
\end{tabular}
\caption{\textbf{Experiment Organization}: An overview of the different properties and experiments for each experimental dataset.}
\label{table:exp_setup}
\end{table*}

\begin{table*}[t]
\centering
\small
\begin{tabular}{| p{\textwidth} |}
 \hline
 \textbf{CoCoTrip Summary - Annotator 1($S^{A \setminus B}_1$): } This is a hotel located in an excellent place. It is perfect to walk to the entertainment district as well as pubs and restaurants. Rooms on the higher floors are excellent as they can provide a great view. The room itself is of a lovely quality and even had a full kitchen, which is incredibly useful. The hotel's complimentary breakfast was really good. \\
 \hline
 \textbf{CoCoTrip Summary - Annotator 1 ($S^{B \setminus A}_1$) - Reference Contrastive: }This hotel is great and pretty amazing, and you would definitely want to return. The hotel is located close to the CN Tower, the hockey Hall of Fame and the Rogers Centre for example. The deluxe room was quite small and didn't have a view at all, but it was pretty clean. There was a lot of noise that could be heard from the room, as well. The hotel's breakfast is in included in the room's price, but a little expensive. \\
 \hline
 \textbf{CoCoTrip Summary - Annotator 2 ($S^{A \setminus B}_2$) - Reference Similar): }This is a great hotel to visit and you will look forward to coming back again and again. The hotel is situated near lots of pubs and restaurants. The hotel room was on a high floor so there was a great view of the CN Tower. The room also had a full kitchen which was ideal. The morning breakfast was complimentary and served really great superb food. \\
 \hline
 \textbf{Paraphrase($S^{A \setminus B}_1$) - Synthetic Contrast: }This hotel is situated in a prime location. It is ideal to take a stroll to the area of amusement as well as bars and eateries. Rooms located on the upper levels are ideal as they offer a stunning view. The room was of an excellent standard and even had a full kitchen, which was extremely useful. The hotel's free breakfast was exceptionally delicious. \\
 \hline
 \textbf{Decomposed$(S^{A \setminus B}_1)$: }This is a hotel. The hotel is located in an excellent place. It is perfect to walk to the entertainment district. It is perfect to walk to pubs and restaurants. Rooms on the higher floors are excellent. They can provide a great view. The room itself is of a lovely quality. The room had a full kitchen, which is incredibly useful. The hotel offers a complimentary breakfast. The breakfast was really good. \\
 \hline
 \textbf{$\neg$ \textbf{Decomposed}$(S^{A \setminus B}_1)$ - Synthetic High Contrast By Negation: } This is not a hotel. The hotel is located in a terrible place. It is not close enough to walk to the entertainment district. It is not close enough to walk to pubs and restaurants. Rooms on the higher floors are terrible. They provide a disagreeable view. The room itself is of an uncomely quality. The room lacked a full kitchen, which was incredibly incovenient. The hotel provides a paid breakfast. The breakfast was egregious.\\
 \hline
\end{tabular}
\caption{\textbf{Data and Pre-Processing Used In Experiments}}
\label{table:experimental_datasets}
\end{table*}

Table~\ref{table:exp_setup} shows the different experimental datasets referred to in our paper and Table~\ref{table:experimental_datasets} provides a selection of examples from each of these datasets.

\subsection{Paraphrasing}
For paraphrasing, we use the text-davinci-003
large language model from OpenAI. The prompt
used for paraphrasing was just 'Paraphrase this' and the input summary. We set \texttt{max\_tokens=512} and \texttt{temperature=0.5}.

\subsection{Negations}
For creating negations, we first split each sentence using the process discussed with \texttt{gpt-3.5-turbo}, since it is non trivial to negate complex-claim sentences. For each single-claim sentence for a summary we obtain negated sentences by way of manual annotation. This achieved by adding modifiers ('not'), substituting complements of quantifiers ('everyone' to 'no one'), or using antonyms ('close' to 'far away'). Then we re-combine negated simple-claim sentences into complex-claim sentences using a rule-based method of simple conjunctions, since our baselines operate on complex-claim sentences.

\section{Calculating $\metric$}
\label{sec:calculating_caspr}

We describe here how we calculate the $\metric$ score after obtaining single-claim sentences by the process described in Appendix~\ref{sec:sentence_decomposition}. We compare each single-claim sentence $s^{A \setminus B}_i$ in summary $S^{A \setminus B}$ with every single-claim sentence $s^{B \setminus A}_j$ in summary $S^{B \setminus A}$. For each comparison $(s^{A \setminus B}_i,s^{B \setminus A}_j)$, we define the indicator function $1_{\text{cont}}$ for the 'CONTRADICTION' label as,
\begin{multline*}
    1_{\text{cont}} = 
    \begin{cases}
        1, & 
    \begin{aligned}[t]
        \text{NLI}(s^{A \setminus B}_i \rightarrow s^{B \setminus A}_j)= \text{cont}\\
        \textbf{and } \text{NLI}(s^{B \setminus A}_j \rightarrow s^{A \setminus B}_i) \neq \text{ent}\\
    \end{aligned}\\
    \\
    1, & 
    \begin{aligned}[t]
        \text{NLI}(s^{B \setminus A}_j\rightarrow s^{A \setminus B}_i)= \text{cont}\\
        \textbf{and } \text{NLI}(s^{A \setminus B}_i \rightarrow s^{B \setminus A}_j) \neq \text{ent}\\
    \end{aligned}\\
    \\
        0, & otherwise
    \end{cases}\\
\end{multline*}
That is, if the NLI label for $(s^{A \setminus B}_i,s^{B \setminus A}_j)$ in either direction is contradiction, and the label in the other direction is not entailment, the indicator function for contradiction has a value of 1. We get the indicator function $1_{\text{ent}}$ of the 'ENTAILMENT' label similarly. For the 'NEUTRAL' indicator function we have,
\begin{multline*}
    1_{\text{neut}} = 
    \begin{cases}
        1, & 
    \begin{aligned}[t]
        \text{NLI}(s^{A \setminus B}_i \rightarrow s^{B \setminus A}_j)= \text{neut}\\
        \textbf{and } \text{NLI}(s^{B \setminus A}_j\rightarrow s^{A \setminus B}_i)= \text{neut}\\
    \end{aligned}\\
    \\
     1, & 
    \begin{aligned}[t]
        \text{NLI}(s^{A \setminus B}_i \rightarrow s^{B \setminus A}_j)= \text{cont}\\
        \textbf{and } \text{NLI}(s^{B \setminus A}_j\rightarrow s^{A \setminus B}_i)= \text{ent}\\
    \end{aligned}\\
    \\
    1, & 
    \begin{aligned}[t]
        \text{NLI}(s^{A \setminus B}_i \rightarrow s^{B \setminus A}_j)= \text{ent}\\
        \textbf{and } \text{NLI}(s^{B \setminus A}_j\rightarrow s^{A \setminus B}_i)= \text{cont}\\
    \end{aligned}\\
    \\
        0, & otherwise
    \end{cases}\\
\end{multline*}
We now aggregate the values of the indicator functions $1_{\text{cont}},1_{\text{ent}},1_{\text{neut}}$ over all sentences $s^{B \setminus A}_j \in S^{B \setminus A}$ for the sentence $s^{A \setminus B}_i$ to get the counts $n^i_{\text{cont}},n^i_{\text{neut}},n^i_{\text{neut}}$ For example, we get $n^i_{\text{cont}}$ as follows,
\[
    n^i_{\text{cont}} = \sum_{s^{B \setminus A}_j \in S^{B \setminus A}} 1_{\text{cont}} (s^{A \setminus B}_i, s^{B \setminus A}_j)
\]
We then get the label score $l(s^{A \setminus B}_i|S^{B \setminus A})$ of the sentence by comparing the aggregates in the following way,
\begin{equation*}
l(s^{A \setminus B}_i|S^{B \setminus A})  =
    \begin{cases}
        +1, & n^i_{\text{neut}} = |S^{B \setminus A}| \\
        \\
        -1, & n^i_{\text{ent}} >= n^i_{\text{cont}} \\
        \\
        +1, & otherwise\\
    \end{cases}
\end{equation*}
where $n^i_{\text{neut}} = |S^{B \setminus A}|$ denotes the case where all comparisons evaluate to 'neut' in both directions.
We them sum the label scores across all sentences of both summaries.
\begin{multline*}
    L_{AB} = \sum\limits_{s^{A \setminus B}_i \in S^{A \setminus B}} l(s^{A \setminus B}_i|S^{B \setminus A})\\
    + \sum\limits_{s^{B \setminus A}_i \in S^{B \setminus A}}
    l(s^{B \setminus A}_i|S^{A \setminus B})
\end{multline*}
We then normalize $L_{AB}$ across the total number of sentences across both summaries.
\[
 \hat{L} = \frac{L_{AB}}{|S^{A \setminus B}| + |S^{B \setminus A}|}
\]
$\hat{L}$ lies in the range $[-1,1]$. We finally rescale it to lie in the range $[0,100]$.
\[
 \text{\metric} = \frac{\hat{L} + 1}{2} * 100
\]


\begin{figure*}[t]
\includegraphics[width=\textwidth]{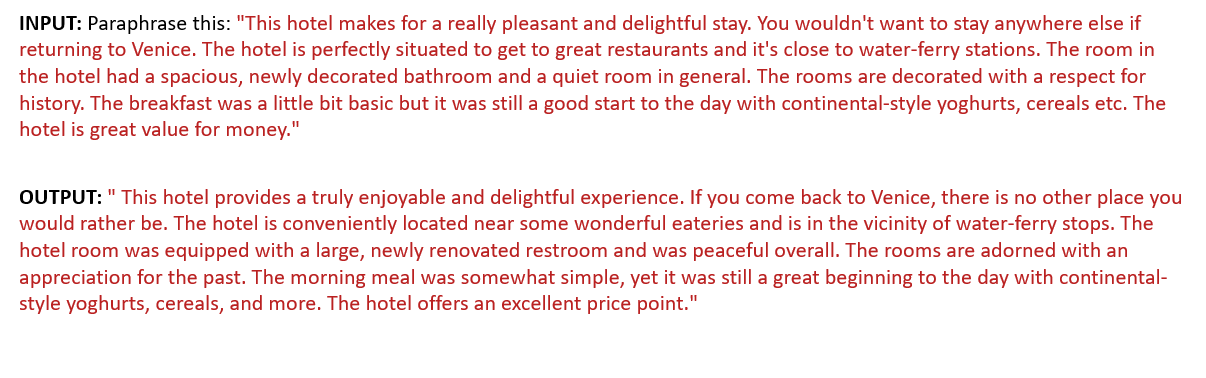}
\caption{Paraphrase example with prompt input 'Paraphrase this' using \texttt{text-davinci-003}}
\label{fig:paraphrase_example_prompt}
\end{figure*}

\section{Supplementary Results: Paraphrasing}

   \par \noindent $\metric$ is reasonably robust to lexical changes that preserve meaning. Specifically, if we generate the paraphrases of Reference Contrastive summaries $S^{A \setminus B},S^{B \setminus A}$ to obtain Synthetic Contrastive summaries $P(S^{A \setminus B}), P(S^{B \setminus A})$, mean $\metric$ scores shows negligible differences between the two datasets. $BS^{-1}$ and $DS$ show similar differences. $BS^{-1}(86\%)$ outperforms $\metric(66\%)$ and $DS(51\%)$ in regards to rank correlation between scores from these datasets. This is due to the original purpose of BERTScore as a similarly score between two sentences using semantic embeddings. An explanation of this dataset is included in Figure \ref{table:experimental_datasets}.

\begin{table*}[ht]
\centering
\small
\begin{tabular}{c | c | c | c } 

 Metric & $S^{A \setminus B},S^{B \setminus A}$ & $P(S^{A \setminus B}),P(S^{B \setminus A})$  & Rank Correlation \\ 
 \hline
 DS & 73.6  $\pm$ 0.9 & 74.1 $\pm$ 0.8 & 51\% \\ 

 BS$^{-1}$ & 72.5 $\pm$ 1.4 & 73.4 $\pm$ 1.3 & 86\% \\

 CASPR & 84.3 $\pm$ 3.8 & 83.7 $\pm$ 3.6 &  66\% \\

 \hline
\end{tabular}
\caption{\textbf{Synthetic Contrast (Paraphrasing)} Average scores over the \textsc{CoCoTrip} \textit{Reference Contrastive} dataset $(S^{A\setminus B},S^{B \setminus A})$ and the \textit{Synthetic Contrast} dataset $(P(S^{A \setminus B}), P(S^{B \setminus A}))$ with rank correlation comparing the two scores $CASPR$ and $BS^{-1}$ outperform $DS$ with higher correlation. The scores were computed using Spearman Rank Correlation. $BS^{-1}$ outperforms both due to its original task of comparing similarity between candidate and reference sentences. Yet, CASPR having a lower correlation than BS shows that there is room for improvement.}
\label{table:4}
\end{table*}



\section{Error Analysis of NLI for $\metric$}

\par An interesting scenario (Table \ref{table:6})  where NLI fails is when the hypothesis is a common sense maxims. For e.g.  the hypothesis 'This hotel provides a stay' always evaluates to an entailment, no matter the premise (unless the premise explicitly contradicts it with 'this hotel does not provide a stay' which goes against commonsense). This leads to more entailments and low contrast scores on the Reference Contrastive dataset. This might be remedied with a stronger NLI model and a better sentence decomposition model.

\begin{table*}
\centering
\small
\begin{tabular}{c | c | c } 

 \hline
 Premise & Hypothesis & Label \\ 
 \hline
 The shower was excellent. & \textcolor{red}{This hotel has served guests.} & ENTAILMENT\\ 

 There was no refrigerator in the room.
 & \textcolor{red}{This hotel has served guests.} & ENTAILMENT \\

 The staff at this hotel seemed a little bit uninterested. & \textcolor{red}{This hotel provides a stay.} & ENTAILMENT\\

 \hline
\end{tabular}
\caption{Cases where the NLI model fails. These \textcolor{red}{hypotheses} are universally true statements. For most examples, no matter the premise, the NLI label is entailment. These sentences cause the overall NLI score to be low.}
\label{table:6}
\end{table*}
\end{document}